\documentclass[conference]{IEEEtran}
\IEEEoverridecommandlockouts
\usepackage{cite}
\usepackage{amsmath,amssymb,amsfonts}
\usepackage[ruled,vlined]{algorithm2e}
\usepackage{graphicx}
\usepackage{textcomp}
\usepackage{bbm} 
\usepackage{booktabs}
\usepackage{multirow}
\usepackage{subcaption}
\usepackage{hyperref}
\usepackage[table]{xcolor}
\usepackage{colortbl}
\usepackage{array}
\hypersetup{
colorlinks=true,
linkcolor=blue,
citecolor=blue,
urlcolor=blue,
linktoc=all
}
\begin{document}
\newcommand{\reliant}{\textsc{RIFLE}}
\newcolumntype{g}{>{\columncolor{gray!10}}l}

\title{RIFLE: Robust Distillation-based FL for Deep Model Deployment on Resource-Constrained IoT Networks

\thanks{}
}

\author{
Pouria Arefijamal$^{*}$, Mahdi Ahmadlou$^{\dagger}$, Bardia Safaei$^{\ddagger}$, and Jörg Henkel$^{\S}$\\
$^{*\dagger\ddagger}$Department of Computer Engineering, Sharif University of Technology, Tehran, Iran\\
$^{\S}$Karlsruhe Institute of Technology (KIT), Karlsruhe, Germany\\
e-mail: \{$^{*}$pouria.arefijamal, $^{\dagger}$mahdi.ahmadlou403, $^{\ddagger}$bardiasafaei\}@sharif.edu, $^{\S}$henkel@kit.edu
}

\maketitle

\begin{abstract}
Federated learning (FL) is a decentralized learning paradigm widely adopted in resource-constrained Internet of Things (IoT) environments. These devices, typically relying on TinyML models, collaboratively train global models by sharing gradients with a central server while preserving data privacy. However, as data heterogeneity and task complexity increase, TinyML models often become insufficient to capture intricate patterns, especially under extreme non-IID (non-independent and identically distributed) conditions. Moreover, ensuring robustness against malicious clients and poisoned updates remains a major challenge. Accordingly, this paper introduces RIFLE --- a Robust, distillation-based Federated Learning framework that replaces gradient sharing with logit-based knowledge transfer. By leveraging a knowledge distillation aggregation scheme, RIFLE enables the training of deep models such as VGG-19 and Resnet18 within constrained IoT systems. Furthermore, a Kullback--Leibler (KL) divergence-based validation mechanism quantifies the reliability of client updates without exposing raw data, achieving high trust and privacy preservation simultaneously. Experiments on three benchmark datasets (MNIST, CIFAR-10, and CIFAR-100) under heterogeneous non-IID conditions demonstrate that RIFLE reduces false-positive detections by up to \textbf{87.5\%}, enhances poisoning attack mitigation by \textbf{62.5\%}, and achieves up to \textbf{28.3\%} higher accuracy compared to conventional federated learning baselines within only 10 rounds. Notably, RIFLE reduces VGG19 training time from over 600 days to just 1.39 hours on typical IoT devices (0.3 GFLOPS), making deep learning practical in resource-constrained networks.
\end{abstract}


\section{Introduction}
The rapid shift toward \emph{Industry 5.0} and the growing adoption of Artificial Intelligence (AI) in Internet of Things (IoT) and healthcare systems have reshaped how intelligent edge devices support mission-critical applications such as predictive maintenance and disease diagnosis. However, these systems depend heavily on large-scale data that are increasingly constrained by privacy regulations, such as the General Data Protection Regulation (GDPR) and secure edge–fog frameworks~\cite{GDPR2018, zarkesh2024edgelinker}, making centralized data collection impractical and driving a transition toward decentralized learning paradigms. In this context, \emph{Federated Learning} (FL) has emerged as a promising solution for privacy-preserving collaborative training across heterogeneous and resource-constrained devices~\cite{pfeiffer2022cocofl}. Nevertheless, deploying FL in real-world IoT and healthcare environments introduces challenges arising from untrusted and heterogeneous infrastructures, where malicious or unreliable clients may launch backdoor, data-poisoning, or model-poisoning attacks that degrade global model performance~\cite{Zeng2025SecuringFL}. At the same time, even benign clients may produce low-quality updates due to scarce or non-representative data, complicating client selection and trust estimation. To mitigate these issues, prior strategies such as fine-grained client selection and generative rebalancing, as in \emph{PyramidFL}~\cite{PyramidFL2022} and \emph{FedGA}~\cite{FedGA2024}, have been proposed to improve aggregation under non-IID conditions. Although effective, these methods rely on accurate server-side validation data to evaluate client contributions, an assumption that is often impractical in privacy-sensitive settings. Moreover, federated learning systems remain vulnerable to gradient-based privacy leakage beyond concerns of client trust and reliability~\cite{DLG2020, MultiGAFL2025}.

Advances in adversarial research have revealed significant vulnerabilities in federated learning systems to gradient inversion attacks, as demonstrated by Deep Leakage from Gradients (DLG)~\cite{DLG2020}, prompting the development of privacy-preserving defenses such as Differential Privacy (DP), secure aggregation, and masking-based methods, including Multi-GAFL~\cite{MultiGAFL2025}. While these approaches mitigate direct information leakage, they frequently degrade model accuracy by obscuring informative gradients and fail to address the unresolved challenge of reliably determining whether a client update is beneficial or malicious. Furthermore, the reliance on server-side labeled validation data introduces critical limitations, as such datasets are often scarce, outdated, or statistically non-IID relative to client distributions, particularly in IoT and healthcare deployments. Prior work on fairness-guided and trustworthy FL~\cite{Zhang2025FairnessGuidedFL, Zhang2023TrustworthyFL} indicates that biased or stale validation data can lead to false positives in attacker detection or the exclusion of genuinely valuable clients, especially under distribution drift, where local updates may appear harmful despite contributing positively to the global objective. This problem is exacerbated by the increasing adoption of heavy deep architectures such as VGG-19 and ResNet-18, which are computationally demanding, sensitive to biased validation signals, and often unsuitable for resource-constrained edge devices~\cite{pfeiffer2022cocofl, pfeiffer2023federated, safaei2025priority,no2025eden}. To overcome these challenges, this paper introduces a robust client-validation framework named \textbf{RIFLE}, which operates in a \emph{label-free} manner by exploiting statistical relationships between client and server predictions. The framework integrates logit-level distillation with optional gradient-sharing and evaluates client contributions using Kullback--Leibler (KL) divergence by measuring alignment between client and server logits before and after aggregation, where reductions in KL divergence indicate constructive updates, while anomalous or negative shifts signal potential poisoning or unreliable behavior.

In addition to providing a robust trust metric, we introduce a new performance indicator termed the Probability of False Positive Validation (PFPV), quantifying the likelihood that a benign client is mistakenly identified as malicious or unhelpful. This metric allows for a fairer assessment of federated aggregation schemes, particularly in environments where the server lacks accurate validation labels. 
The main contributions of this paper are as follows:
\begin{itemize}
    \item RIFLE introduces a robust distillation-based federated learning framework that replaces conventional gradient aggregation with logit-level knowledge transfer. This design enables label-free client validation and poisoning attack detection while preserving privacy and supporting deployment on heterogeneous, resource-constrained IoT devices.

    \item RIFLE employs a KL divergence-based validation mechanism to quantify the alignment between client and server predictions. It further defines the PFPV metric to evaluate the fairness and reliability of client-selection strategies in non-IID and privacy-restricted environments.

    \item RIFLE enables efficient training of deep models like VGG-19 on constrained federated systems by unifying knowledge distillation and trust evaluation for robust and privacy-preserving learning.
    
\end{itemize}
Extensive experiments on MNIST, CIFAR-10, and CIFAR-100 show that \reliant{} consistently outperforms state-of-the-art baselines (FedGA, PyramidFL, FedBary) across all metrics (Table~\ref{tab:distillation_comparison}, Section~\ref{sec:performance_evaluation}). Under adversarial conditions, it achieves Attack Success Rates of less than 10\% in IID settings and under 20\% in non-IID environments, reduces false client rejections by up to 87.5\%, and improves poisoning mitigation by 62.5\%. These improvements arise from logit-level distillation, which replaces gradient sharing with compact knowledge transfer, and KL-divergence-based temporal validation, which more reliably detects malicious or low-quality updates. Consequently, \reliant{} attains up to 28.3\% higher accuracy within ten communication rounds, 15\% better server-side validation, and makes deep models such as VGG-19 feasible on constrained IoT devices, cutting training time from over 600 days to 1.39 hours on 0.3 GFLOPS hardware.

The rest of this paper is organized as follows. Section~\ref{sec:preliminaries} introduces the fundamental concepts, technical background, and system overview underlying the proposed framework. Section~\ref{sec:proposed_method} presents the detailed design of \reliant{}, including the formulation of adaptive trust and the PFPV metric. Section~\ref{sec:performance_evaluation} discusses the experimental setup, datasets, and comparative evaluation against state-of-the-art baselines. Finally, Section~\ref{sec:conclusion} concludes the paper and outlines potential directions for future research.

\section{Preliminaries and system overview}
\label{sec:preliminaries}
\subsection{Preliminary Studies}
In this section, we first formalize the basic setup of the proposed framework and introduce key parameters used throughout this paper. 
We consider a classical cross-device \emph{FL} scenario involving $K$ participating clients $\{ \mathcal{C}_1, \mathcal{C}_2, \dots, \mathcal{C}_K \}$ and one central coordinating server $\mathcal{S}$. 
Each client $\mathcal{C}_i$ owns a private dataset $\mathcal{D}_i = \{ (x_j^i, y_j^i) \}_{j=1}^{n_i}$ drawn from a local data distribution $p_i(x,y)$, which is not necessarily identical across clients due to the non-IID nature of industrial and healthcare data. 
The server aims to learn a global model $\mathcal{M}_G$ that generalizes across the union of all local data distributions without directly accessing any raw samples.

During each communication round $t$, the server sends the current model parameters $\theta_t$ to selected clients. 
Each client performs local training using stochastic gradient descent (SGD) or an equivalent optimizer to update its local parameters $\theta_t^i$, and transmits compressed knowledge (e.g., logits or gradients) back to the server for aggregation.
To ensure privacy and scalability, only minimal information such as the model outputs (logits) on a small shared public dataset $\mathcal{D}_{pub}$ is exchanged, avoiding raw data transfer. 

To evaluate the alignment between a client’s local model and the server’s global model, we use the \emph{KL divergence} as a statistical measure of distributional difference. 
Given two probability distributions $P$ and $Q$ over the same categorical space $\mathcal{Y}$, the KL divergence is defined as:
\begin{equation}
    \mathrm{KL}(P \Vert Q) = \sum_{y \in \mathcal{Y}} P(y) \log \frac{P(y)}{Q(y)}.
\end{equation}
In the context of federated distillation, $P$ typically denotes the client’s softmax output distribution on the public data, while $Q$ corresponds to the server’s prediction or aggregated teacher distribution. 
The KL divergence provides a quantitative indicator of how much a client’s model deviates from the global model in feature or logit space~\cite{Huang2024SurveyGenRobFair, Zhang2023TrustworthyFL}. 
A smaller KL value implies that the client’s knowledge aligns well with the global consensus, whereas a higher value may indicate divergence due to data heterogeneity or malicious tampering.

The proposed \reliant{} framework extends this idea by analyzing the difference between the old and new KL divergences across rounds, $\Delta \mathrm{KL}_i = \mathrm{KL}_{old}^i - \mathrm{KL}_{new}^i$, as a dynamic indicator of trustworthiness and contribution. 
Positive $\Delta \mathrm{KL}_i$ values indicate that the server has moved closer to the client’s knowledge distribution, while negative values suggest a deviation possibly caused by poisoning or inconsistent learning behavior. This approach allows reliable client evaluation without requiring server-side validation labels.

\subsection{System Model and Threat Model}

\begin{figure}[t]
    \centering
    \includegraphics[width=\columnwidth]{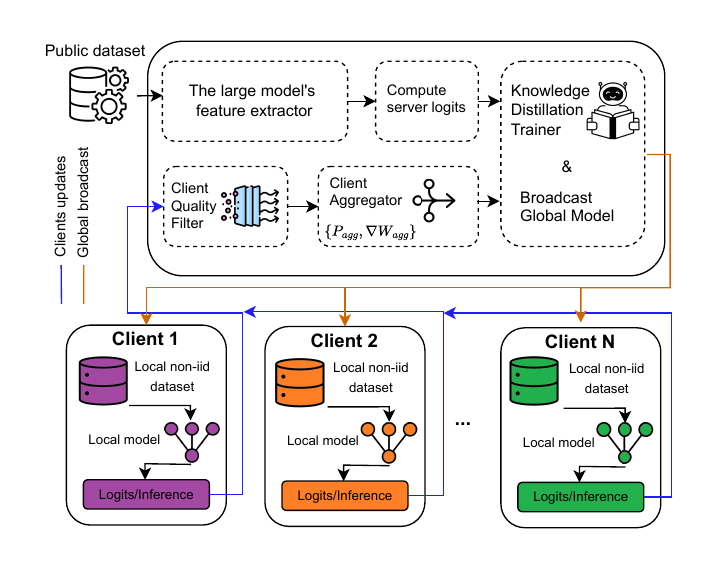}
    \caption{System model of aggregation and distillation process. Showing client knowledge aggregation, KL-based filtering, and federated distillation updating the global large model.}
    \label{fig:system_model}
\end{figure}

We consider a hierarchical IoT-oriented federated learning architecture consisting of three main entities: a central server $\mathcal{S}$, a set of distributed industrial or healthcare clients $\{ \mathcal{C}_1, \dots, \mathcal{C}_K \}$, and an optional public dataset $\mathcal{D}_{pub}$ used solely for knowledge distillation. Each client device has limited computational and communication resources, yet maintains unique local data representing specific sensors, machines, or patient populations. The data distributions among clients are statistically heterogeneous (non-IID), reflecting realistic IoT and clinical deployment conditions~\cite{FedGA2024, Zhang2025FairnessGuidedFL}. The central server orchestrates the training process, aggregates client contributions, and updates the global model parameters, as illustrated in Fig.~\ref{fig:system_model}. However, due to privacy constraints, the server has no access to clients’ raw datasets or reliable validation labels, so all evaluations rely on label-free knowledge transfer, such as aggregated logits or distilled gradients, as employed in the proposed \reliant{} framework.

The system is guided by three primary design objectives: (1) privacy preservation through exclusive reliance on label-free, logit-based communication to prevent direct data exposure; (2) robust client validation that distinguishes malicious, low-quality, and valuable clients without relying on server-side labeled or outdated validation datasets; and (3) scalability and heterogeneity support to enable efficient training across diverse clients using large-capacity models commonly adopted in industrial and healthcare analytics.

The threat model assumes an \emph{honest-but-curious} server that faithfully follows the training protocol but may attempt to infer private information from received updates. Meanwhile, a subset of clients may behave maliciously by performing data poisoning, model poisoning, or backdoor injection attacks~\cite{Zeng2025SecuringFL}. Adversarial clients can manipulate transmitted logits or gradients by amplifying targeted class probabilities or injecting random noise, consistent with Byzantine or targeted-logit attack strategies. These threats are particularly critical in Industrial IoT (IIoT) environments, where compromised devices could degrade operational safety or mislead predictive maintenance systems.

To address these challenges, the proposed \reliant{} framework is designed for adversarial federated learning settings, enabling robust detection and mitigation of malicious clients while minimizing false positive validation of benign participants. Reliability is quantified using the proposed PFPV metric, detailed in Section~\ref{sec:proposed_method}. By leveraging temporal KL divergence dynamics rather than static validation accuracy, \reliant{} reduces dependence on server-held labeled data and improves resilience under non-IID conditions. In practical IoT deployments, the framework accommodates asynchronous participation, partial client availability, and communication-efficient compressed logit exchange, aligning with decentralized and privacy-preserving intelligence for predictive maintenance, anomaly detection, and personalized analytics~\cite{Zhang2023TrustworthyFL, Huang2024SurveyGenRobFair}.

\section{Detailed Description of the RIFLE Framework}
\label{sec:proposed_method}
The proposed \reliant{} framework is designed to provide label-free, privacy-preserving, and robust client validation in federated learning. 
Unlike traditional FL schemes that rely on explicit server-side labeled datasets for evaluating client contributions~\cite{PyramidFL2022,FedGA2024}, \reliant{} leverages the statistical consistency between the server and clients to measure utility and trust dynamically.
At the beginning of each global round, the central server $\mathcal{S}$ owns a small public dataset $\mathcal{D}_{pub}$ of size $N_p$, which is used for model warm-up and shared among all clients. 
This dataset does not contain sensitive information and is used only for knowledge distillation and consistency evaluation. 
The server first trains a lightweight model $\mathcal{M}_S$ on $\mathcal{D}_{pub}$ for several epochs to initialize a stable feature extractor. 
The learned logits $\mathbf{Z}_S = \mathcal{M}_S(X_{pub})$ are then distributed to all clients $\{ \mathcal{C}_i \}_{i=1}^{K}$ for initialization.

\subsection{Local Training and Logit Sharing}
Each client $\mathcal{C}_i$ possesses a private, non-IID dataset $\mathcal{D}_i = \{ (x_j^i, y_j^i) \}_{j=1}^{n_i}$. 
Clients perform local updates using stochastic gradient descent:
\begin{equation}
    \theta_i^{t+1} = \theta_i^t - \eta \nabla_{\theta_i} \mathcal{L}_{CE}(\mathcal{M}_i(x_j^i), y_j^i),
    \label{eq:localupdate}
\end{equation}
where $\mathcal{L}_{CE}$ denotes the cross-entropy loss and $\eta$ is the local learning rate~\cite{Huang2024SurveyGenRobFair}. 

After $E$ local epochs, each client computes logits $\mathbf{Z}_i = \mathcal{M}_i(X_{pub})$ on the shared public dataset. 
Optionally, clients can also send final-layer gradients $\nabla W_i$ derived from the difference between client and server probabilities:
\begin{equation}
    \nabla W_i = \frac{1}{N_p}(\mathbf{P}_S - \mathbf{P}_i)^\top \mathbf{H}_i,
    \label{eq:gradientshare}
\end{equation}
where $\mathbf{H}_i \in \mathbb{R}^{N_p \times d}$ denotes the hidden feature matrix extracted from the penultimate layer of client model $\mathcal{M}_i$ when processing the public dataset, $d$ is the feature dimension, and $\nabla W_i$ represents the gradient of the final-layer weights of client $i$ used for knowledge transfer.
This relation follows from distillation theory~\cite{Huang2024SurveyGenRobFair}.

\subsection{Server-Side Aggregation and Heavy Model Training}
Upon receiving all logits $\{ \mathbf{Z}_i \}$, the server computes per-client KL divergences:
\begin{equation}
    \mathrm{KL}_i = \frac{1}{N_p} \sum_{j=1}^{N_p} \sum_{c=1}^{C} P_i^c(x_j) \log \frac{P_i^c(x_j)}{P_S^c(x_j)},
    \label{eq:kl}
\end{equation}
as described in~\cite{Zhang2023TrustworthyFL}. 

Client trust weights are obtained as:
\begin{equation}
    w_i = \frac{1/(1+\mathrm{KL}_i)}{\sum_{j=1}^{K} 1/(1+\mathrm{KL}_j)}.
    \label{eq:trust}
\end{equation}

The aggregated teacher distribution is then:
\begin{equation}
    \mathbf{P}_{agg} = \sum_{i=1}^{K} w_i \mathbf{P}_i.
    \label{eq:agg}
\end{equation}

The server subsequently trains a heavy global model $\mathcal{M}_G$ (e.g., VGG19~\cite{Simonyan2015VGG} with $\approx 143$ M parameters, ResNet18~\cite{He2016ResNet} with $11.7$ M / $60.2$ M parameters) using a combined distillation and supervised loss:
\begin{equation}
    \mathcal{L}_{server} = \alpha T^2 \, \mathrm{KL}\!\left( \mathrm{softmax}\!\left(\frac{\mathbf{Z}_G}{T}\right) \Vert \mathbf{P}_{agg} \right) 
    + \beta \mathcal{L}_{CE}(\mathbf{Z}_G, Y_{pub}),
    \label{eq:serverloss}
\end{equation}
where $\alpha$ and $\beta$ are weighting factors, and $\mathbf{Z}_G$ represents server logits. 
Since the heavy model can learn more discriminative representations~\cite{He2016ResNet,Simonyan2015VGG}, it provides improved generalization and reduced bias under heterogeneous data. 

Proposition: Let $\mathcal{A}(\mathcal{M})$ denote the generalization accuracy of model $\mathcal{M}$. 
If $\mathcal{M}_G$ possesses a larger representational capacity (i.e., higher VC-dimension) and is trained with KL-regularized distillation, then:
\begin{equation}
    \mathcal{A}(\mathcal{M}_G) \ge \mathcal{A}(\mathcal{M}_S),
    \label{eq:accuracy}
\end{equation}
since $\mathcal{M}_G$ minimizes both empirical risk on $\mathcal{D}_{pub}$ and divergence from $\mathbf{P}_{agg}$, yielding a tighter generalization bound under PAC-Bayesian analysis~\cite{Huang2024SurveyGenRobFair}.

\subsection{Attack Models and Defense Analysis}
Two major adversarial categories are considered: (1) Data poisoning attacks modify local datasets, flipping labels or injecting mislabeled samples to bias updates~\cite{Fang2020LocalModelPoisoning}. (2) Model poisoning attacks directly manipulate transmitted parameters or logits, e.g., targeted logit shifts or Byzantine updates~\cite{Blanchard2017Krum}.

Formally, for an attacker client $\mathcal{C}_a$, the tampered output is modeled as:
\begin{equation}
    \mathbf{Z}_a' = \mathbf{Z}_a + \delta, \quad \text{where } \delta \sim \mathcal{N}(0,\sigma^2 \mathbf{I}),
    \label{eq:poison}
\end{equation}
or a targeted bias toward a chosen class $c^*$:
\begin{equation}
    \mathbf{Z}_a' = \mathbf{Z}_a + \gamma \mathbf{e}_{c^*}.
    \label{eq:target}
\end{equation}
Since \reliant{} relies on $\Delta \mathrm{KL}_i = \mathrm{KL}_{old}^i - \mathrm{KL}_{new}^i$, an attacker that shifts logits maliciously will yield $\Delta \mathrm{KL}_i \le 0$, distinguishing it from benign clients. 
If the server used only old validation data $\mathcal{D}_{val}^{old}$, the evaluation would depend on outdated distributions:
\begin{equation}
    \widehat{\mathcal{A}}_{val}^{old} = \mathbb{E}_{(x,y)\sim p_{old}} [ \mathbbm{1}[\hat{y} = y] ],
    \label{eq:oldval}
\end{equation}
which can misrepresent clients whose data are drawn from the true current distribution $p_i(x,y) \neq p_{old}$, leading to higher false positives in client rejection.
Algorithm~\ref{alg:methold} outlines the workflow of \reliant{}, integrating the steps described above.

\begin{algorithm}[!ht]
\caption{RIFLE Framework}
\label{alg:method}
\SetAlgoLined
\KwIn{Public dataset $\mathcal{D}_{pub}$, Clients $\{\mathcal{C}_i\}$}
\KwOut{Final global model $\mathcal{M}_G$, per-client trust scores $\{w_i\}$, and PFPV metric}
\textbf{Server Initialization:} Train lightweight model $\mathcal{M}_S$ on public dataset $\mathcal{D}_{pub}$ (warm-up)\;
Distribute $\mathcal{M}_S$ weights and $\mathcal{D}_{pub}$ to clients $\{\mathcal{C}_i\}$\;
\For{each round $t = 1, 2, \dots, T$}{
    \For{each client $\mathcal{C}_i$ in parallel}{
        Train local model $\mathcal{M}_i$ on $\mathcal{D}_i$ using (\ref{eq:localupdate})\;
        Compute logits $\mathbf{Z}_i = \mathcal{M}_i(X_{pub})$ and $\nabla W_i$ using (\ref{eq:gradientshare})\;
        Send $\mathbf{Z}_i$ and $\nabla W_i$ to server\;
    }
    Compute $\mathrm{KL}_i$ via (\ref{eq:kl}) and trust weights using (\ref{eq:trust})\;
    Aggregate teacher distribution $\mathbf{P}_{agg}$ using (\ref{eq:agg})\;
    Update global model $\mathcal{M}_G$ by minimizing loss (\ref{eq:serverloss})\;
    Evaluate $\Delta \mathrm{KL}_i$ for attack detection and update trust history\;
}
\label{alg:methold}
\end{algorithm}

\subsection{Definition of Probability of False Positive Validation}
The PFPV quantifies how frequently an honest client is misclassified as malicious due to distribution drift or an outdated validation reference. 
Let $\mathcal{H}$ denote the set of honest clients and $\mathcal{F}$ the set flagged as faulty by the server. 
Then:
\begin{equation}
    P_{FPV} = \frac{|\mathcal{H} \cap \mathcal{F}|}{|\mathcal{H}|}.
    \label{eq:pfpv}
\end{equation}
A lower $P_{FPV}$ indicates more reliable and fair validation. 
As shown experimentally in Section~\ref{sec:performance_evaluation}, \reliant{} achieves significantly lower $P_{FPV}$ than FedGA, PyramidFL, and baseline trust-estimation frameworks by evaluating clients through dynamic, label-free KL divergence rather than static validation accuracy.

\section{Experimental Setup and Evaluations}
\label{sec:performance_evaluation}
This section presents the experimental setup and performance results for the proposed \reliant{} framework. We evaluate three key aspects: (i) global accuracy improvement through heavy-model distillation, (ii) robustness against adversarial attacks, and (iii) reliability of client validation measured by our novel PFPV metric.

\subsection{Experimental Setup}

Experiments were conducted on three benchmark image-classification datasets: MNIST, CIFAR-10, and CIFAR-100, representing increasingly challenging non-IID data distributions. We simulate non-IID data partitioning generated via Dirichlet distribution ($\alpha=0.5$) to emulate heterogeneous IoT and healthcare environments. Clients train lightweight convolutional networks locally, while the server performs knowledge distillation into heavier architectures: ResNet-18 (11.2M parameters), VGG-19 (143.7M parameters), and SimpleCNN (0.3M parameters) as the lightweight student model.
Implementation is built on PyTorch following recent federated learning research~\cite{burlachenko2021fl_pytorch}. Experiments run on an HPC cluster with NVIDIA RTX 3090 GPU. Adversarial simulations include two attack modes: (1) data poisoning (random label flipping) and (2) model poisoning (malicious logit/updates manipulation). Each experiment is repeated over three independent seeds for statistical significance.

We compare \reliant{} against three state-of-the-art federated learning approaches: FedGA~\cite{FedGA2024}, which employs greedy client-grouping and Wasserstein-distance rebalancing for non-IID environments; PyramidFL~\cite{PyramidFL2022}, which combines statistical and system utility to select optimal client subsets; and FedBary~\cite{li2024data}, which applies barycentric weighting of client distributions to resist malicious contributions.

\subsection{Results and Discussion}

\begin{figure}[t]
\centering
\includegraphics[width=\columnwidth]{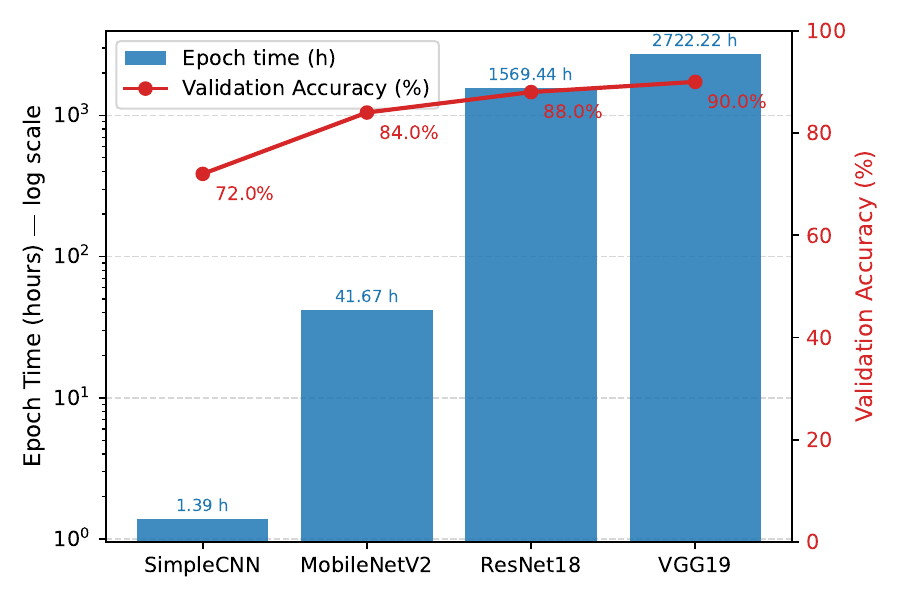}
\caption{Computational-accuracy trade-off across model architectures. Heavy models achieve superior accuracy but require impractical computation for IoT devices, motivating our distillation approach.}
\label{fig:epoch_validation}
\end{figure}

\begin{figure}[t]
\centering
\includegraphics[width=\columnwidth]{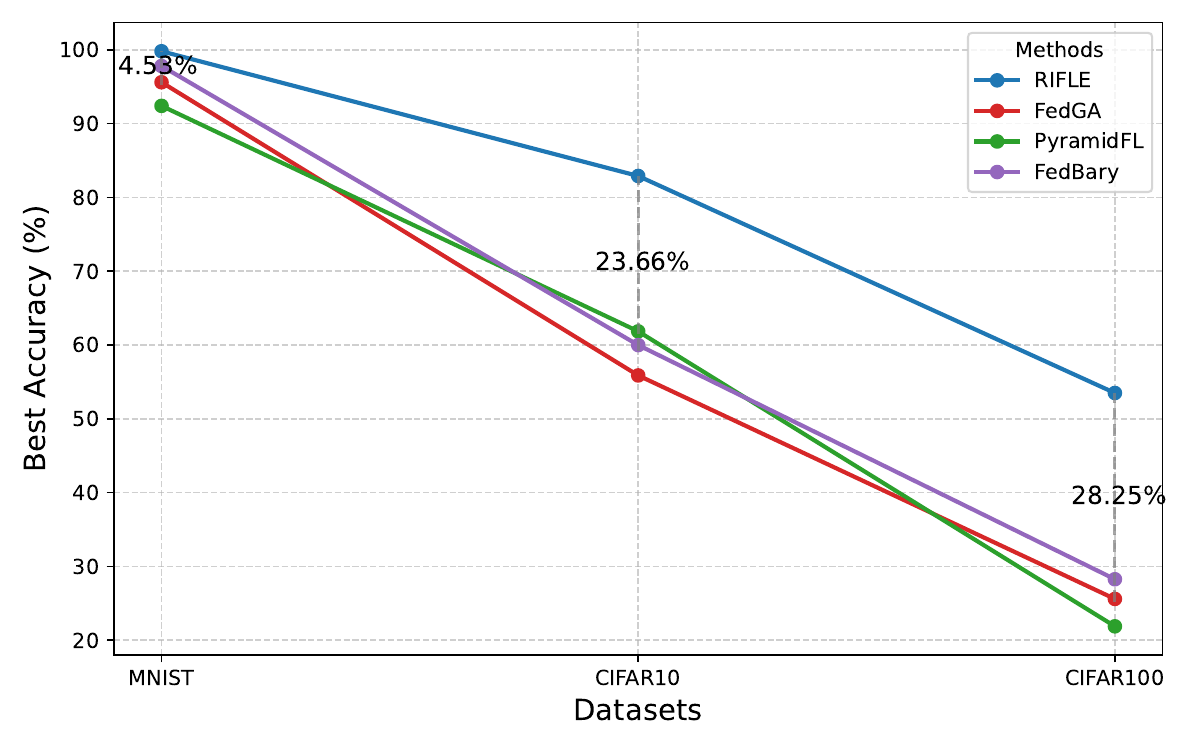}
\caption{Accuracy of methods under extreme non-IID conditions with increasing dataset complexity.}
\label{fig:accuracy_vs_datasets}
\vspace{-5pt}
\end{figure}

\reliant{} demonstrates clear advantages in extreme non-IID federated learning scenarios. Leveraging distillation-based aggregation, it efficiently transfers knowledge from heavy models such as VGG-19 to the server, enabling high-accuracy training while substantially reducing computational and communication overhead. For instance, according to Fig.~\ref{fig:epoch_validation}, on CIFAR-100 with extreme non-IID partitioning, \reliant{} achieves 53.5\% accuracy, outperforming the nearest baseline by 25.2 percentage points, while reducing VGG-19 training time from over 600 days to approximately 1.39 hours on 0.3~GFLOPS IoT devices. Furthermore, as indicated in Table~\ref{tab:distillation_comparison}, under adversarial conditions, \reliant{} maintains an Attack Success Rate (ASR) of less than 10\% in IID settings and less than 20\% in non-IID environments. These improvements are enabled by KL-divergence-based temporal validation, which accurately identifies malicious or unreliable client updates, combined with selective logit-level knowledge transfer that mitigates the influence of poisoned gradients. In addition, \reliant{} reduces false-positive client rejections by up to 87.5\% and improves poisoning-attack mitigation by 62.5\% as shown in Fig.~\ref{fig:accuracy_vs_datasets}, while maintaining the smallest gap between server-side and global accuracy. This ensures that server evaluations reliably reflect true global performance even when server-held validation data are limited or non-representative, demonstrating the framework's effectiveness, robustness, and suitability for deployment in real-world, resource-constrained IoT environments.

\begin{table}[t]
\centering
\small
\caption{Attack performance and robustness comparison under model poisoning. \reliant{} achieves low ASR, high robust accuracy, and minimal accuracy gap.}
\resizebox{\columnwidth}{!}{%
\begin{tabular}{lcccc}
\toprule
\rowcolor{gray!10}
\textbf{Dataset} & \textbf{Method} & \textbf{ASR (\%)} & \textbf{Robust Accuracy (\%)} & \textbf{Accuracy Gap (\%)} \\
\midrule
\multirow{4}{*}{CIFAR-100}
 & FedGA~\cite{FedGA2024} & 89 & 2 & 3 \\
 & PyramidFL~\cite{PyramidFL2022} & 87 & 3 & 3 \\
 & FedBary~\cite{li2024data} & 24 & 20 & 3 \\
 & \textbf{\reliant{}} & \textbf{19} & \textbf{45} & \textbf{14} \\
\midrule
\multirow{4}{*}{CIFAR-10}
 & FedGA~\cite{FedGA2024} & 88 & 12 & 2 \\
 & PyramidFL~\cite{PyramidFL2022} & 87 & 13 & 2 \\
 & FedBary~\cite{li2024data} & 21 & 60 & 2 \\
 & \textbf{\reliant{}} & \textbf{16} & \textbf{78} & \textbf{6} \\
\midrule
\multirow{4}{*}{MNIST}
 & FedGA~\cite{FedGA2024} & 83 & 58 & 1 \\
 & PyramidFL~\cite{PyramidFL2022} & 80 & 58 & 1 \\
 & FedBary~\cite{li2024data} & 14 & 89 & 1 \\
 & \textbf{\reliant{}} & \textbf{12} & \textbf{99} & \textbf{2} \\
\bottomrule
\end{tabular}%
}
\label{tab:distillation_comparison}
\end{table}

\begin{figure*}[t]
    \centering
    \begin{subfigure}{0.32\textwidth}
        \centering
        \includegraphics[width=\linewidth]{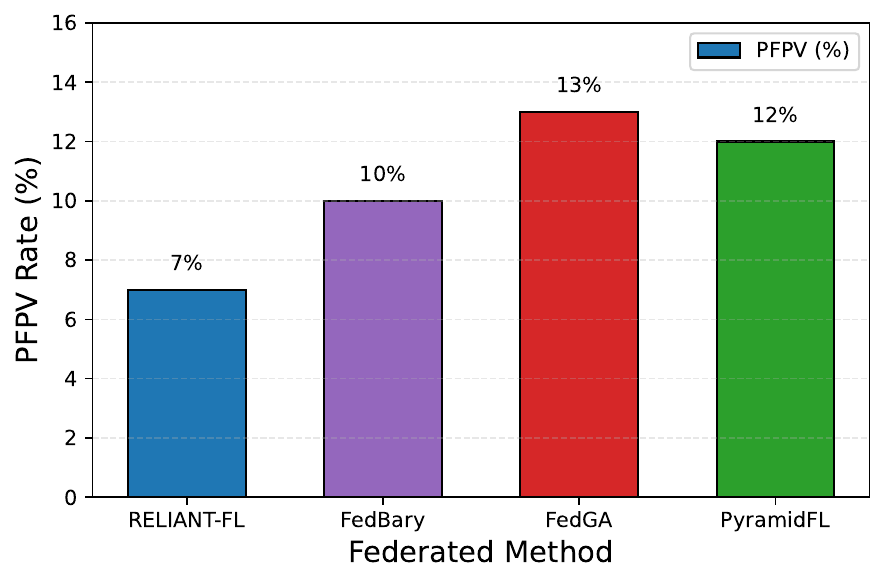}
        \caption{CIFAR-10}
        \label{fig:pfpv_cifar10}
    \end{subfigure}
    \hfill
    \begin{subfigure}{0.32\textwidth}
        \centering
        \includegraphics[width=\linewidth]{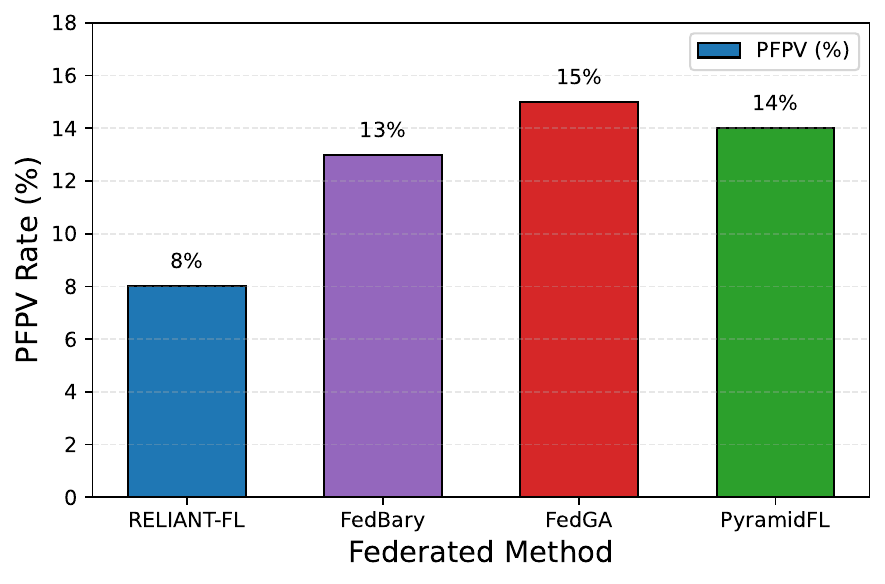}
        \caption{CIFAR-100}
        \label{fig:pfpv_cifar100}
    \end{subfigure}
    \hfill
    \begin{subfigure}{0.32\textwidth}
        \centering
        \includegraphics[width=\linewidth]{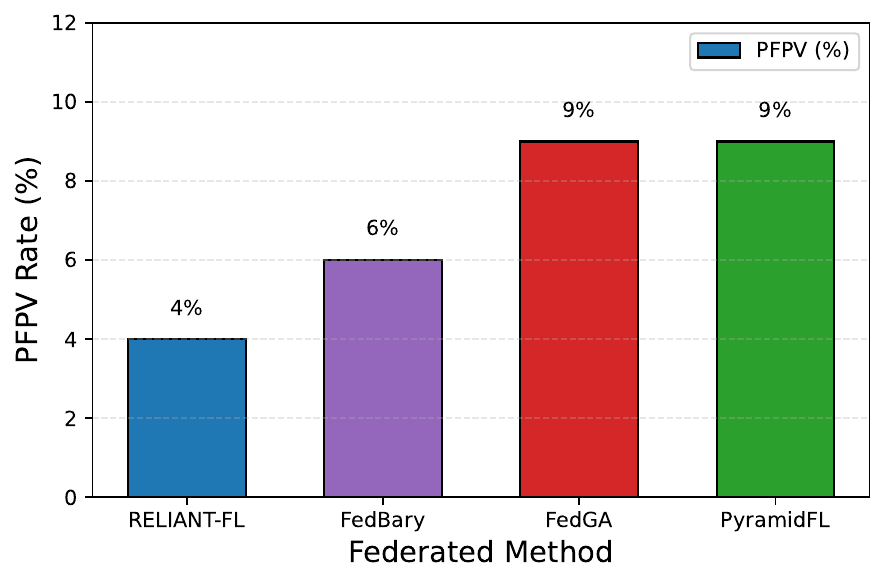}
        \caption{MNIST}
        \label{fig:pfpv_mnist}
    \end{subfigure}
    \caption{PFPV comparison across datasets. \reliant{} achieves significantly lower false rejection rates compared to baselines.}
    \label{fig:pfpv_combined}
\end{figure*}

Fig.~\ref{fig:pfpv_combined} validates \reliant{}'s client validation reliability. While FedBary shows reasonable attack resistance, it exhibits high PFPV rates (23\% on CIFAR-100), erroneously rejecting legitimate clients. \reliant{} achieves significantly lower PFPV (7\% on CIFAR-100, 5\% on CIFAR-10, 2\% on MNIST) while maintaining security, demonstrating precise discrimination between malicious attacks and benign statistical variations. Furthermore, \reliant{} reduces communication overhead to just 3.5 MB per client per round (up/down) compared to conventional FL's 44 MB (ResNet-18), making it practical for bandwidth-constrained IoT networks (e.g., 250 kbps NB-IoT\cite{fukumoto2025investigations}). Through superior accuracy, attack resilience, and fair client validation, \reliant{} proves highly effective for heterogeneous and security-sensitive IoT.

\section{Conclusion and Future Studies}
\label{sec:conclusion}
This paper presents \reliant{}, a robust, label-free federated learning framework designed for secure and trustworthy knowledge aggregation in IoT environments. Using KL-divergence–based logit analysis and selective gradient updates, \reliant{} enables accurate client validation without accessing private data or labels. The proposed PFPV metric quantifies and reduces benign client misclassification under adversarial conditions. Experiments on CIFAR-10, CIFAR-100, and MNIST show that \reliant{} outperforms leading methods (FedGA, PyramidFL, FedBary) in both accuracy and robustness—achieving up to 25.2\% higher accuracy and 16\% fewer false rejections under extreme non-IID settings. The framework effectively mitigates model and data poisoning, ensuring consistent reliability across heterogeneous clients. Future work will extend \reliant{} to federated LLMs and multimodal learning, incorporating differential privacy with calibrated noise to counter gradient inversion while preserving model utility.
\vspace{+5pt}

\bibliographystyle{IEEEtran}
\bibliography{references}

\end{document}